\documentclass[letterpaper, 10 pt, conference]{ieeeconf}  
\IEEEoverridecommandlockouts                                                           
\usepackage{cite}
\usepackage{amsmath}
\usepackage{graphicx}
\usepackage[font=footnotesize,labelfont=bf,skip=0pt]{caption}
\usepackage{epstopdf}
\usepackage{amssymb}
\usepackage{enumerate}
\usepackage{etaremune}
\usepackage{setspace}
\usepackage{comment}
\usepackage{color}

\title{\LARGE \bf
Vehicle-Human Interactive Behaviors in Emergency: Data Extraction from Traffic Accident Videos
}
\author{Wansong Liu$^1$, Danyang Luo$^2$, Changxu Wu$^3$, Minghui Zheng$^{4,*}$
\thanks{$^1$Wansong Liu is with the Department of Mechanical and Aerospace Engineering, University at Buffalo, Buffalo, NY 14260, USA. Email: \tt\small{wansongl@buffalo.edu.}}
\thanks{$^2$Danyang Luo is with the Department of Computer Science and Engineering, University at Buffalo, Buffalo, NY 14260, USA. Email: \tt\small{dluo2@buffalo.edu.}}
\thanks{$^3$Changxu Wu is with the Department of Systems and Industrial Engineering, The University of Arizona, AZ 85721, USA. Email: \tt\small{changxuwu@email.arizona.edu.}}
\thanks{$^{4}$Minghui Zheng is with the Department of Mechanical and Aerospace Engineering, University at Buffalo, Buffalo, NY 14260, USA. Email: \tt\small{mhzheng@buffalo.edu.}}
\thanks{$^*$Corresponding Author.}
}
\begin{document}
\maketitle
\thispagestyle{empty}
\pagestyle{empty}
%%%%%%%%%%%%%%%%%%%%%%%%%%%%%%%%%%%%%%%%%%%%%%%%%%%%%%%%%%%%%%%%%%%%%%%%%%%%%%%%
\begin{abstract}
Currently, studying the vehicle-human interactive behavior in the emergency needs a large amount of datasets in the actual emergent situations that are almost unavailable. Existing public data sources on autonomous vehicles (AVs) mainly focus either on the normal driving scenarios or on emergency situations without human involvement. To fill this gap and facilitate related research, this paper provides a new yet convenient way to extract the interactive behavior data (i.e., the trajectories of vehicles and humans) from actual accident videos that were captured by both the surveillance cameras and driving recorders. The main challenge for data extraction from real-time accident video lies in the fact that the recording cameras are un-calibrated and the angles of surveillance are unknown.
The approach proposed in this paper employs image processing to obtain a new perspective which is different from the original video's perspective. Meanwhile, we manually detect and mark object feature points in each image frame. In order to acquire a gradient of reference ratios, a geometric model is implemented in the analysis of reference pixel value, and the feature points are then scaled to the object trajectory based on the gradient of ratios. The generated trajectories not only restore the object movements completely but also reflect changes in vehicle velocity and rotation based on the feature points distributions.
\end{abstract}
%%%%%%%%%%%%%%%%%%%%%%%%%%%%%%%%%%%%%%%%%%%%%%%%%%%%%%%%%%%%%%%%%%%%%%%%%%%%%%%%

\section{Introduction}
\noindent
\textcolor{black}{While autonomous vehicles (AVs) have been considered with huge potential in road safety, with the rapid development of AVs and related techniques in autonomous driving \cite{zhou2017hierarchical}, accidents of AVs involving human (i.e., pedestrians, pedal cyclists, and motorcyclist) have received great attention and become a major concern for the acceptance of AVs by the public.}
A statistic data from Pedestrian Traffic Fatalities by State shows that the pedestrian fatalities in 2018 have increased more than $50\%$ over 2009 \cite{[1]}. 
\textcolor{black}{From the accident prevention point of view, the last five seconds before an accident happens is the most safety-critical time period in which movement of vehicles and human can be co-adapted to minimize the chance of injuries and fatalities. The study \textcolor{black}{of} the vehicle-human interactive behavior in the emergency are extensively dependent on the behavior data in emergency situations, which is currently almost unavailable.}

\textcolor{black}{As a key input to the human-vehicle interaction research, extensive studies have been conducted to detect the motion of both human and vehicles recently.} To name a few, Zhao et al. \cite{[2]} combined Histogram of Oriented Gradients (HOG) detection and particle filter to track pedestrians in dynamic backgrounds. Haritaoglu et al. \cite{[3]} created a subsystem based on shape analysis of video imagery, the subsystem detects and tracks models of pedestrians under complex environment. Zhang et al. \cite{zhang2019longitudinal} built up High-angle Spatial-Temporal Diagram Analysis (HASDA) model to transfer the pixel coordinates in traffic surveillance videos to the real-world coordinates, and extracted a vehicle’s trajectory. Coifman et al. \cite{coifman1998real} utilized image processing to locate corner features of vehicles, and created a detection model to extract the vehicle features trajectory when a vehicle is moving in a certain detection region. However, only a few kinds of researches focus on trajectory extraction from real-world traffic accident videos, especially between pedestrians and cars. General methods for pedestrian and car trajectory extraction present several limitations. For example: a detection module can only be utilized on pedestrian motion capture \cite{[3]}; trajectory generation relies on multiple cameras \cite{lipton1998moving}; and the gap exists between real-world traffic accidents and conflicts' reconstruction in simulation \cite{duan2017driver}. 

\begin{figure}[!htbp]
	\centering 
	\includegraphics[scale=0.38]{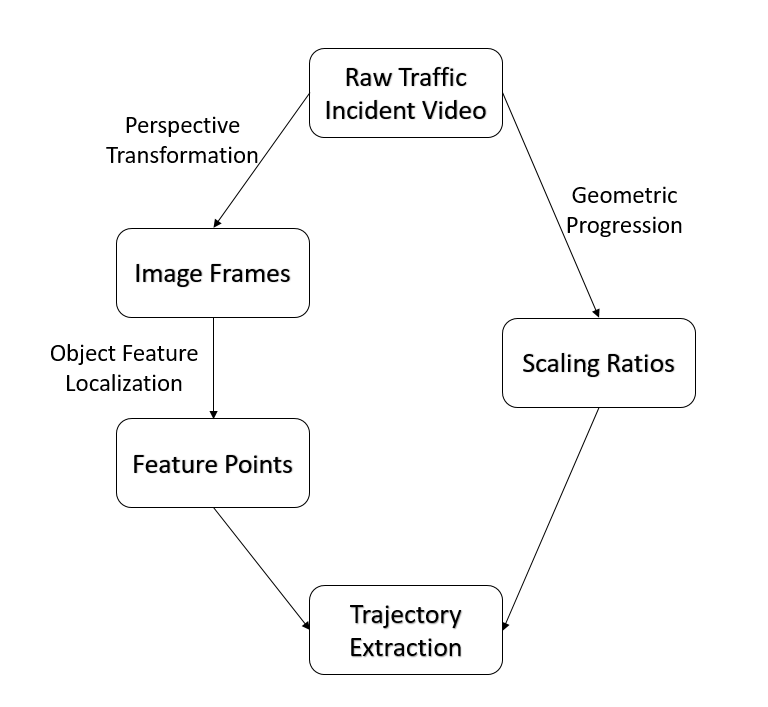}
	\caption{The Architecture of Framework}	\label{fig:Framework}
	\vspace{-20pt}
\end{figure}

\textcolor{black}{Existing public data sources on AVs \cite{aguiar2007trajectory, chee1994lane, funke2016collision, delp2016autonomous} mainly focus either on the normal driving scenarios or under emergency situations without involving human, which is unsuitable for the study of the vehicle-human interactive behaviors in emergency.} This paper aims to present an easy method to extract pedestrian and vehicle trajectory from real-world traffic accident videos. The framework includes the image processing part and the geometric theory part.
The block diagram in Fig.~\ref{fig:Framework} shows the architecture of the framework. 
The lack of efficient video processing approach prompts the transformation between raw traffic incident videos and several pictures. To reduce the difference between raw videos and pictures, unorganized pictures become a continuous image sequence ground on the constant time intervals between each image frame.  Some condensed feature points indicate the continuous movement of the object on the image frames. For the purpose of trajectory reconstruction, operators use a geometric model to obtain scale ratio, and transfer condensed feature point to the object’s real-world position points.

 Image processing has been employed in many areas including transportation and infrastructural health monitoring \cite{liang2019image} as well as traffic incident prediction. Hu et al. \cite{hu2004traffic} created a probabilistic three-dimensional model to predict traffic incidents by employing computer vision and image processing. Zou et al. \cite{zou2009image} combined image processing and hidden Markov model (HMM) to propose a traffic incident detection method. \textcolor{black}{However, these methods neglect the angle between surveillance camera and ground in raw traffic incident videos, resulting in inaccurate reconstruction of object movement. In order to reduce such errors, the method in this paper splits raw video into several image frames, and image processing is utilized to obtain image frames with a new perspective.}  Then, objects detection is an inevitable process \cite{hsieh2006automatic,gupte2002detection,beymerreal}. The factors to a lower accuracy rate of object detection are the resolution of the camera, shadows, posture, congestion, etc. 
 
 To deal with these issues, we mark the feature of an object manually in each image frame. Afterward, based on the pixel value of every maker in the image sequence, feature points of the object are extracted. On the other hand, the length of an object in raw video and in the real world can generate a ratio. This paper assumes a relationship between ratios of points along the road is geometric progression. After that, object trajectory is extracted by applying the length of an object in the real world as a reference to scale feature points. Considering this method calculates the distance between pixel points automatically, and reduces error during vehicle detection, thus the efficiency and accuracy of the trajectory extraction are the emphases.  

\textcolor{black}{Using the proposed method, two types of data sets have been preliminarily collected in this paper. (1) Firstly, we collected some online accident videos that were captured by the surveillance cameras. These cameras have unknown angles with the ground, which plague the analysis of the speed and yaw angle of the moving objects in the videos. We first used the perspective transformation method to recover the unknown camera angles, and then generated a series of images from the videos. By marking the human and the vehicles in each image, their trajectories were extracted after calibration. (2) The second data set was collected by the research team based on online accident videos that were captured by driving recorders. Such videos usually recorded the interactive behaviors (e.g., horning and/or lighting of the vehicle, the VRUs’ response, and the vehicle’s maneuver). This data set includes the trajectories of the vehicle and the human, and reveals the warning actions taken by the vehicle and the subconscious reactions taken by the pedestrian. }
\vspace{-5pt}
\section{Image Processing}
\vspace{-5pt}
\subsection{Affine Transformation}
\vspace{-5pt}
An affine transformation is a linear transformation from one set of two-dimensional coordinates to another, while a straight line remains straight after the transformation with unchanged relative positional relationship of images. Any affine transformation could be represented as a linear transformation plus a vector. For example, 
\begin{equation}
\begin{small}
\left[ \begin{array}{l}{x^{\prime}} \\ {y^{\prime}} \\ {1}\end{array}\right]{=}\left[ \begin{array}{ccc}{m_{11}} & {m_{12}} & {m_{13}} \\ {m_{21}} & {m_{22}} & {m_{23}} \\ {0} & {0} & {1}\end{array}\right] \left[ \begin{array}{l}{x} \\ {y} \\ {1}\end{array}\right]\triangleq M \left[ \begin{array}{l}{x} \\ {y} \\ {1}\end{array}\right]
\end{small}
\end{equation}
which maps the point ($x$, $y$) to ($x'$, $y'$) by using a 2$\times$3 matrix. The $M$ matrix is a combination of linear transformation and translation, and the elements of $M$, i.e., $m_{11}$, $m_{12}$, $m_{21}$, and $m_{22}$, are the linear variation parameters. $m_{13}$ and $m_{23}$ are translation parameters. The last row of $M$ is fixed with 0, 0, 1, to reduce the matrix from the 3$\times$3 dimensional to  2$\times$3 dimension.

\begin{figure}[ht]
\vspace{-10pt}
	\centering 
	\includegraphics[scale=0.3]{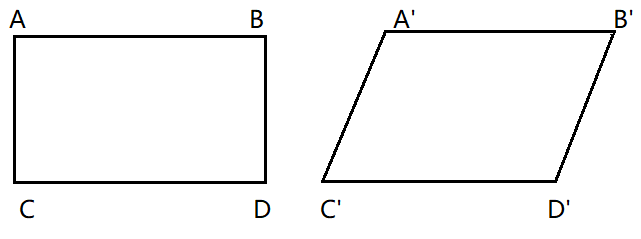}
	\caption{An Affine Transformation Illustration} \label{fig:affine}
		\vspace{-10pt}
\end{figure}
Fig.~\ref{fig:affine} shows the mapping from point $A$ to $A’$, $B$ to $B’$, $C$ to $C’$ and $D$ to $D’$. By implementing the affine transformation, current coordinate will be transformed into a new coordinate with the same dimension. What’s more, the lines are still parallel to each other, the parallelism of lines are constant.

\subsection{Perspective Transformation}
Considering traffic accident videos online have various surveillance angles, the image processing needs a more flexible approach. In an affine transformation, the relative position invariance of lines turns into a limitation in car accident video analysis. On the other hand, perspective transformation, the evolution of the affine transformation, maps an image to a new view plane by projecting this image from a two-dimensional space to a three-dimensional space and then projecting the image in the three-dimensional space into the original two-dimensional space. Compared to the affine transformation, the perspective transformation provides a more flexible way to change the image view angle. It converts an arbitrary quadrilateral region to another one whose lines do not have to be parallel. Instead of just doing a linear transformation, the matrix also provides another parameter line to accomplish perspective transformation comparing to the affine transformation. The following equations represent the perspective transformation: 

\begin{equation}\label{eq:PT1}
\left[ \begin{array}{l}{X} \\ {Y} \\ {Z}\end{array}\right]=\left[ \begin{array}{lll}{m_{11}} & {m_{12}} & {m_{13}} \\ {m_{21}} & {m_{22}} & {m_{23}} \\ {m_{31}} & {m_{32}} & {m_{33}}\end{array}\right] \left[ \begin{array}{l}{x} \\ {y} \\ {1}\end{array}\right]
\end{equation}

\begin{equation}\label{eq:PF2}
x^{\prime}=\frac{X}{Z}=\frac{m_{11}x+m_{12}y+m_{13}}{m_{31}x+m_{32}y+m_{33}}
\end{equation}

\begin{equation}\label{eq:PF3}
y^{\prime}=\frac{Y}{Z}=\frac{m_{21}x+m_{22}y+m_{23}}{m_{31}x+m_{32}y+m_{33}}
\end{equation}
which indicate the two-step transformation, i.e., (1) from a two-dimensional $(x,y)$ with a $z$ axis value of $1$ to a three-dimensional $(X,Y,Z)$ via a three-by-three transformation matrix, as described in Eq.~(\ref{eq:PT1}), and (2) from $(X,Y,Z)$ to $(x',y')$ by dividing the value on $z$ axis in the three-dimensional plane, as described by Eqs.~(\ref{eq:PF2}) and (\ref{eq:PF3}). In brief, the perspective transformation turns points from 2D plane to 3D plane, and then back to the previous 2D space (instead of another 2D space).

\begin{figure}[ht]
	\centering 
	\includegraphics[scale=0.2]{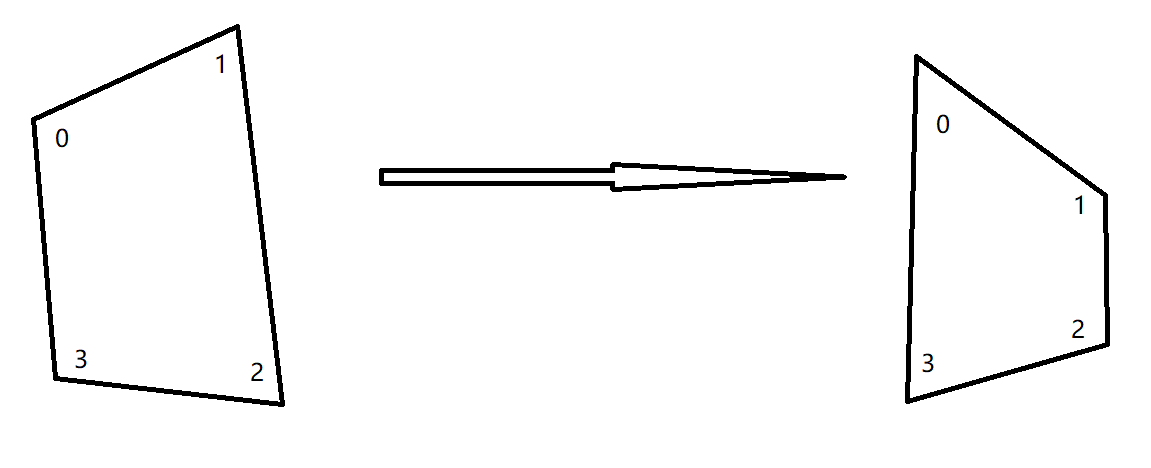}
	\caption{A Perspective Transformation Illustration.} \label{fig:perspective}
\end{figure}

Fig.~\ref{fig:perspective} illustrates the process of the perspective transformation. Firstly, we characterize the corners of an object which forms a quadrilateral. The values of the corner points from the original two-dimensional plane together with $1$ compose three-dimensional vectors. Secondly, perspective transformation is implemented by multiplication of the transformation matrix and four newly formed three-dimensional coordinate vectors. Eventually, four target corner points ($x'$, $y'$) on a two-dimensional space are calculated by the division of points' value on the three-dimensional coordinate.  
\begin{figure}[!htbp]
	\centering 
	\includegraphics[scale=0.3]{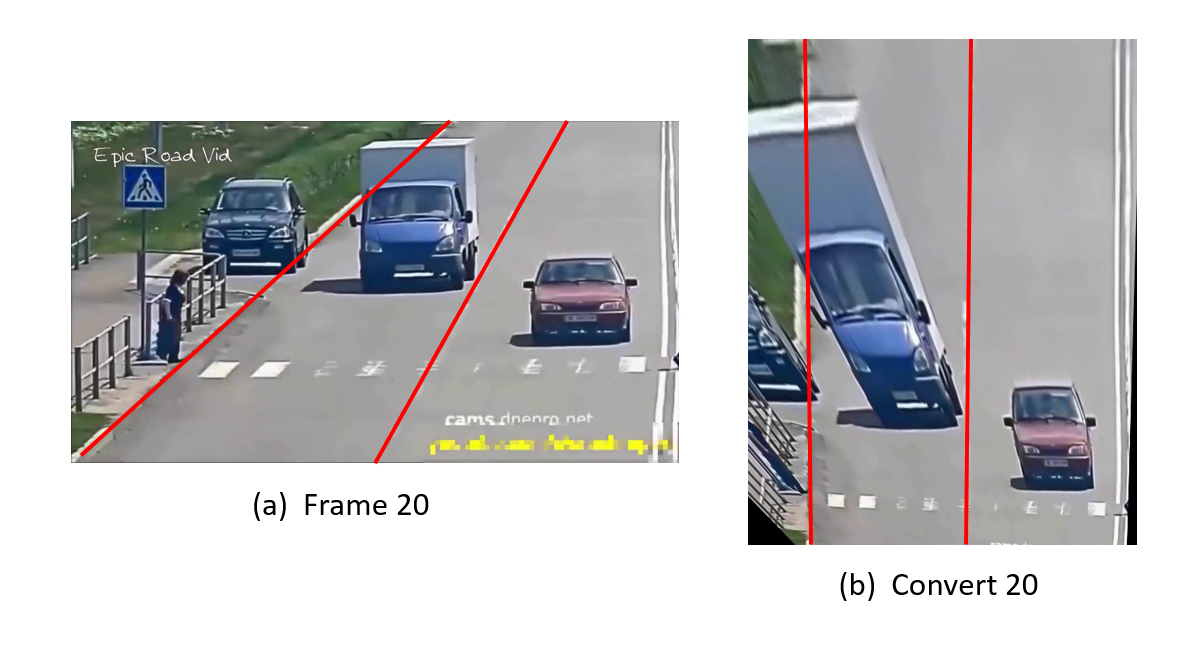}
	\caption{Comparison for Different Perspectives} \label{fig:comparison}
\end{figure}
\vspace{-10pt}

As we know, even though the lane shapes in traffic accident videos are irregular, one physical criterion is that the real-world lines of the lanes are straight and parallel to each other. Therefore, we implemented the perspective transformation to every frame of the videos we collected on-line to obtain a top-down view of the scenario. One example of the perspective transformation for the actual accident video is provided in Fig.~\ref{fig:comparison}.  \textcolor{black}{The picture $(a)$} in Fig.~\ref{fig:comparison} shows the original two-dimensional image and \textcolor{black}{the picture $(b)$} shows the one after perspective transformation. It is worth noted that, in the original image sequence, the vehicle movement is a diagonal line to the picture edge due to the camera angle, while the vehicle trajectory in the converted image sequence is parallel to the edge. A series of object feature points are marked on these converted image frames upon the object movement.

\section{\textcolor{black}{Feature Points Scaling Algorithms}}
\subsection{Surveillance Camera in Street}
The extracted feature points from image sequence are not accurate enough for revealing an object's trajectory in a traffic incident. Therefore, this paper proposes an approach to scale these feature points in the longitudinal and lateral direction separately.

\subsubsection{Longitudinal Direction}
\begin{figure}[!htbp]
	\centering 
	\includegraphics[scale=0.15]{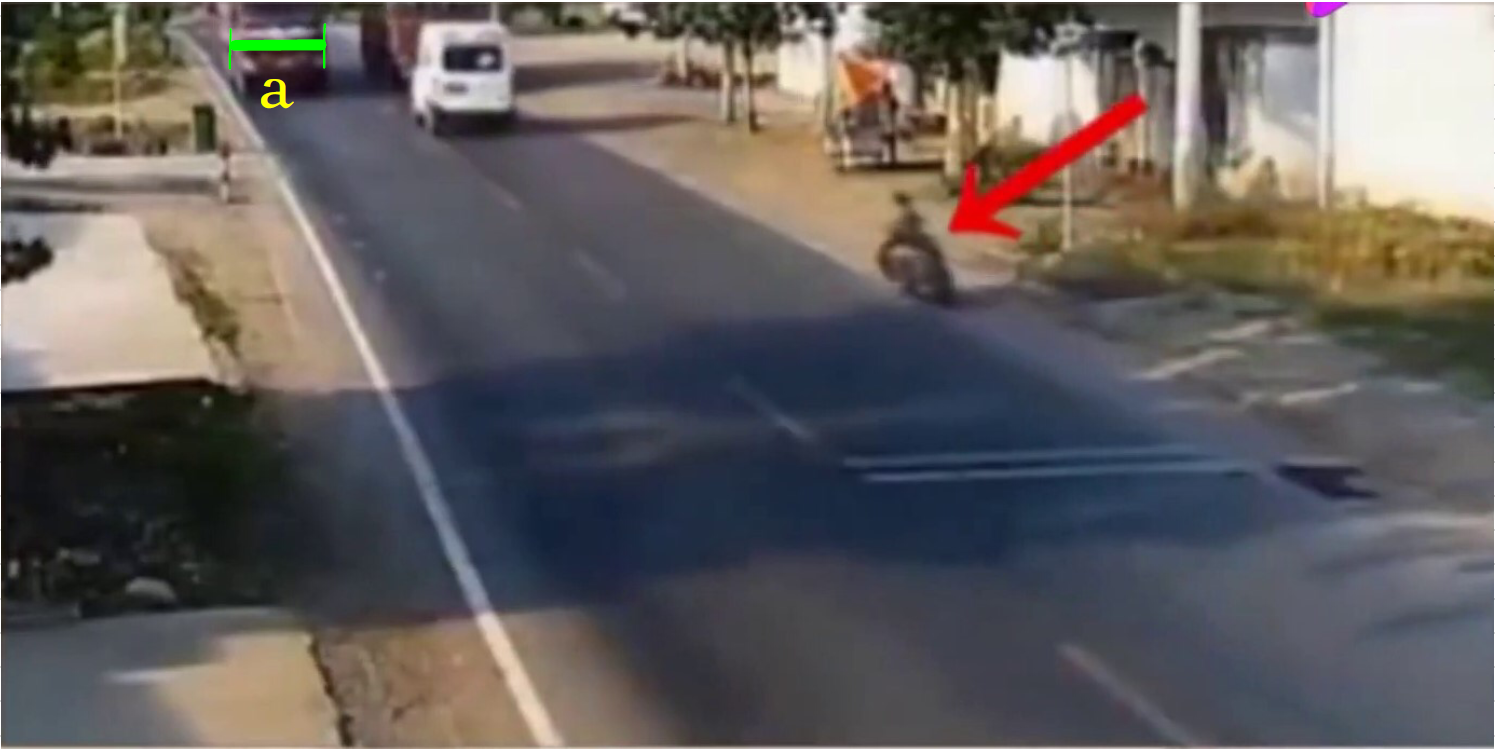}
		\vspace{5pt}
	\caption{The Vehicle's Pixel Width in The First Frame} \label{fig:frame12}
\end{figure}

\vspace{-20pt}
\begin{figure}[!htbp]
	\centering 
	\includegraphics[scale=0.15]{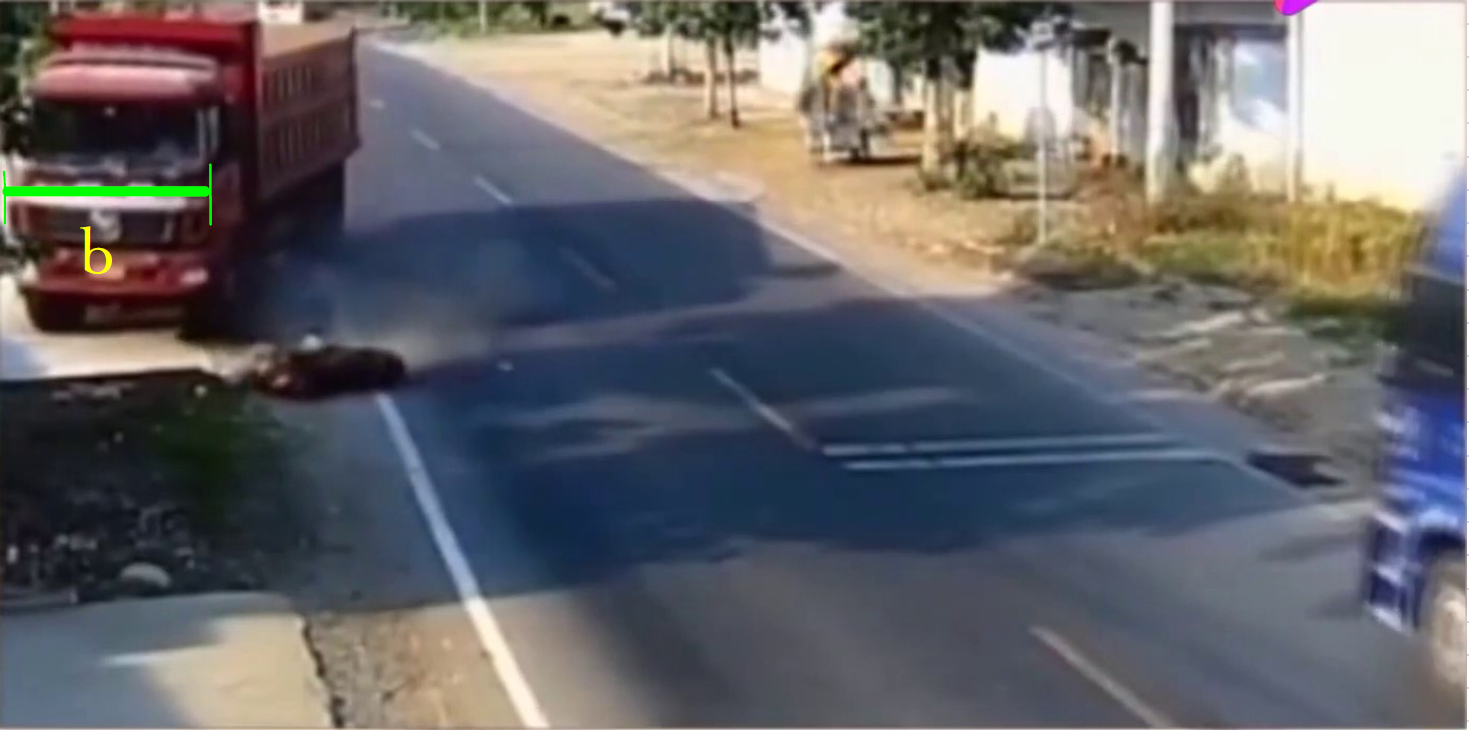}
	\vspace{5pt}
	\caption{The Vehicle's Pixel Width in The Last Frame} \label{fig:frame168}
\end{figure}

\begin{figure}[!htbp]
	\centering 
	\includegraphics[scale=0.15]{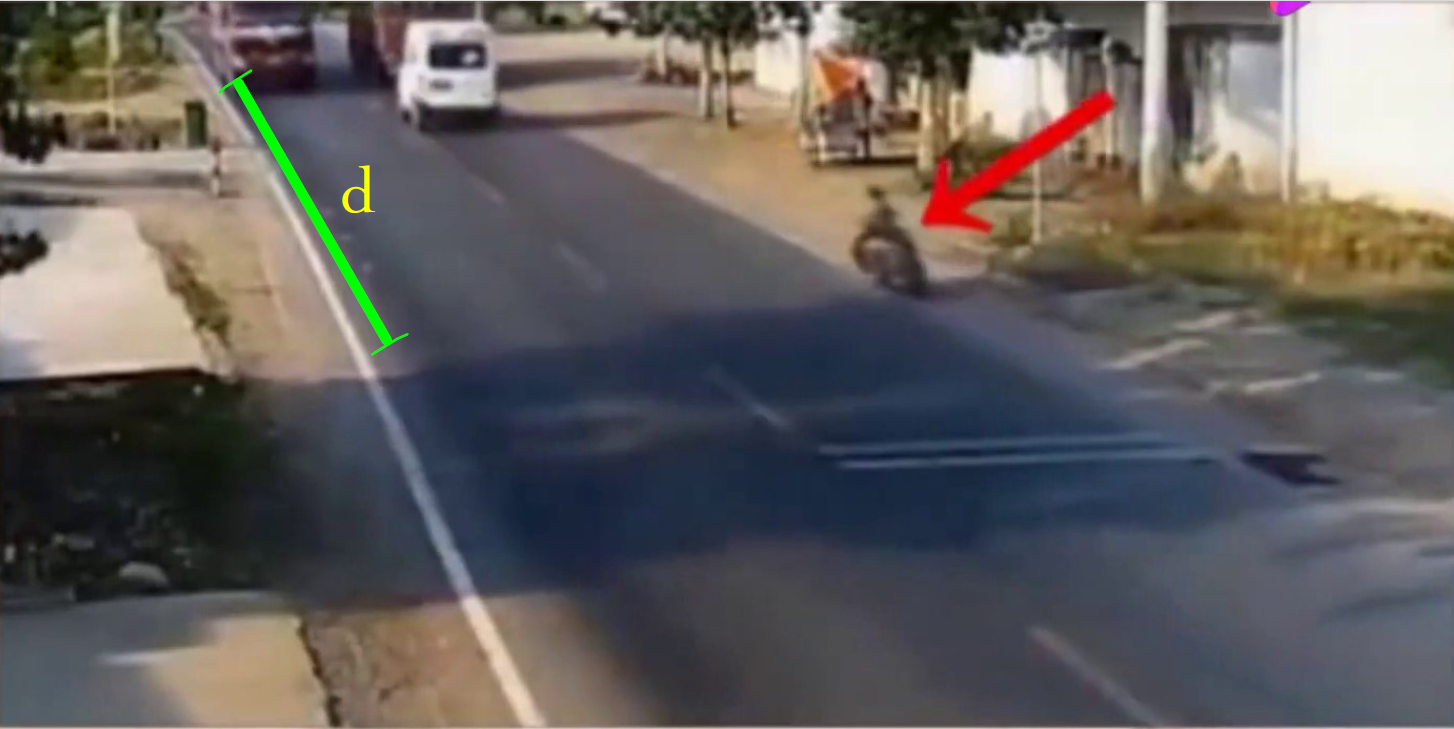}
	\vspace{5pt}
	\caption{Longitudinal Car Movement Distance} \label{fig:movement}
\end{figure}

 Fig.~\ref{fig:frame12} and Fig.~\ref{fig:frame168} depict a truck in different positions. The reference length in pictures is extracted first based on the pixel values. Then it is compared with the real-world reference length to get the ratios for each image frame. By assuming that the ratio gradients in the longitudinal direction are similar to the one of a geometric progression model, this paper uses the following model to represent the ratio:
\begin{equation}
q=\sqrt[i-1]{\frac{r_{i}}{r_{1}}}      \qquad i=1, 2, \cdots, N
\end{equation}
where $N$ is the image frames' total number, $r_{i}$ is ratio in each image frame, respectively, and $q$ is the gradient between each ratio. Fig.~\ref{fig:movement} indicates the truck movement distance in the longitudinal direction in the image sequence. 
Considering that the time intervals between every two consecutive images are the same and the longitudinal vehicle movement is divided into several parts, the vehicle's real-world longitudinal displacement, denoted as $D$, is calculated by the following equation:
\begin{equation}
D\left(y_{i}, h(i), W\right)=\sum_{i=1}^{N}\left(y_{i+1}-y_{i}\right) \frac{W}{h(i)}    \qquad i=1, 2, \cdots, N
\end{equation}
where $W$ is the vehicle's width in the real world, $h(i)$ is the vehicle's width in each image frame, respectively, and $y_{i}$ is the original longitudinal pixel value from feature points. $D$ is employed as a reference to convert the scaled longitudinal pixel points to the real-world longitudinal points. The scaling of feature points in the longitudinal direction is defined as follows:
\begin{equation}
Y_{i+1}=M(D)  \left(Y_{i}+\left(y_{i+1}-y_{i}\right) q^{i-1}\right)    \qquad i=1, 2, \cdots, N
\end{equation}
where $M(D)$ is a transformation function from the pixel value to the real-world length. $Y_{i}$'s are the real-world longitudinal position points of the object after scaling.

\subsubsection{Lateral Direction}
In traffic accidents, pedestrian movement trajectory is usually lateral and perpendicular to the road. Regarding this, this paper assumes that the ratios in the lateral direction are constant.
\begin{figure}[!htbp]
	\centering 
	\includegraphics[scale=0.15]{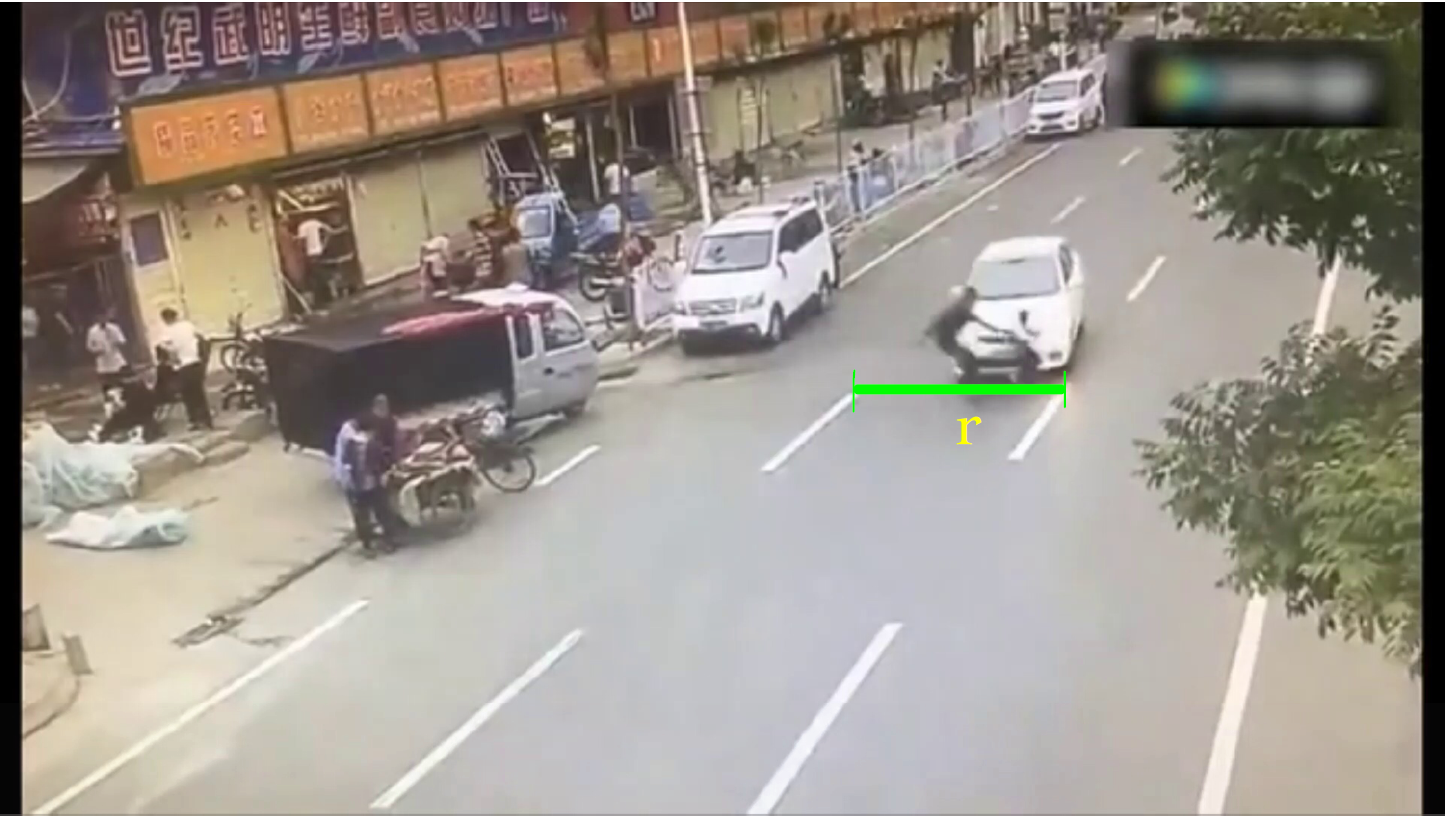}
	\vspace{5pt}
	\caption{Reference in Lateral Direction} \label{fig:reference}
\end{figure}
\vspace{-10pt}
\begin{figure}[!htbp]
	\centering 
	\includegraphics[scale=0.15]{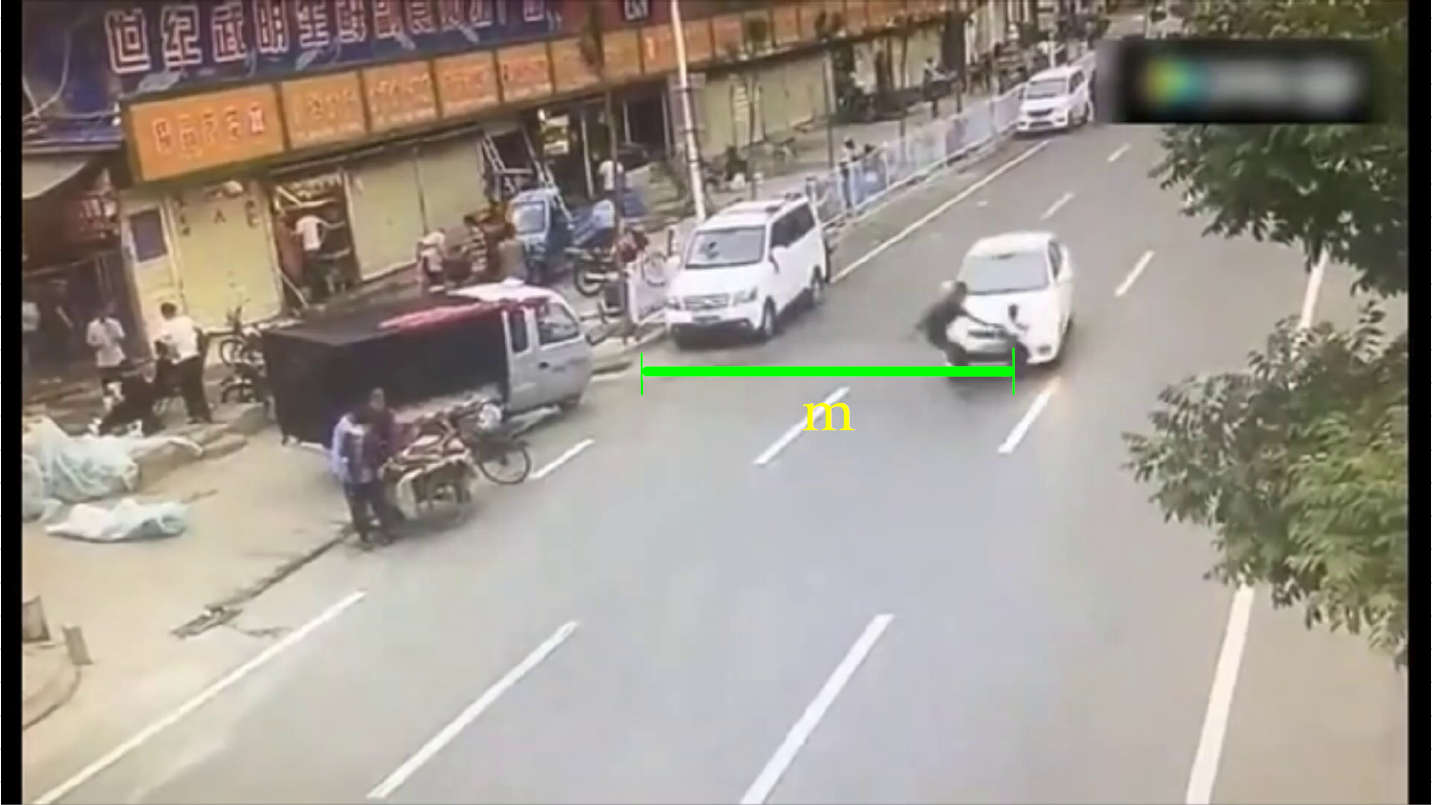}
	\vspace{5pt}
	\caption{Pedestrian Lateral Movement } \label{fig:pedestrian movement}
\end{figure}
 Fig.~\ref{fig:reference} shows the road width which is defined as a lateral reference. A ratio in the lateral direction is obtained by comparing the road width in the image to the one in the real world. The movement of the pedestrian in the image sequence is shown in Fig.~\ref{fig:pedestrian movement}, in which the pedestrian movement pixel values are converted to the real-world distance based on the lateral ratio.

\begin{figure}[!htbp]
	\centering 
	\includegraphics[scale=0.35]{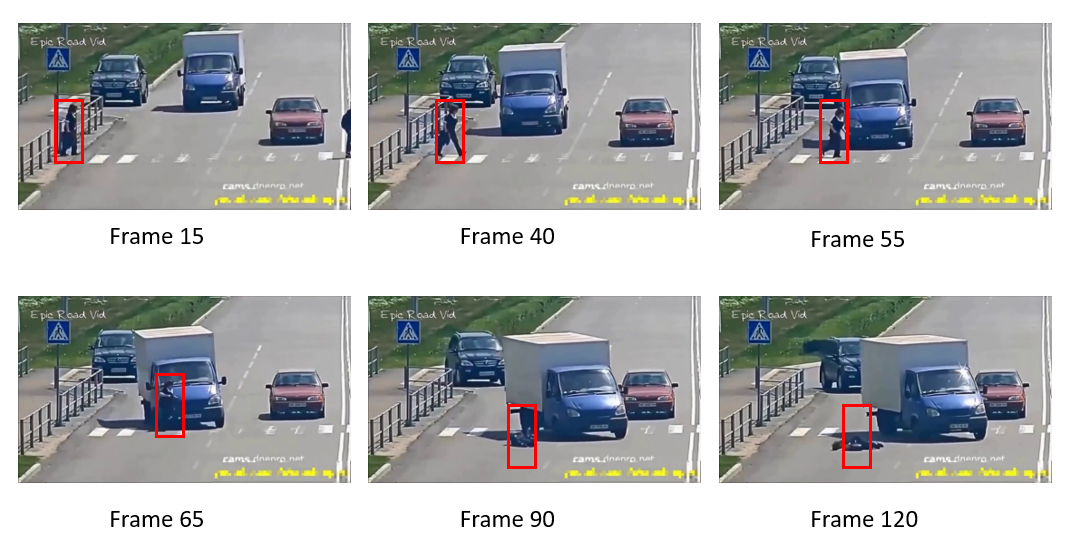} 
	\caption{Original Frames in Video 1} \label{fig:video1}
\end{figure}

\begin{figure}[!htbp]
	\centering 
	\includegraphics[scale=0.35]{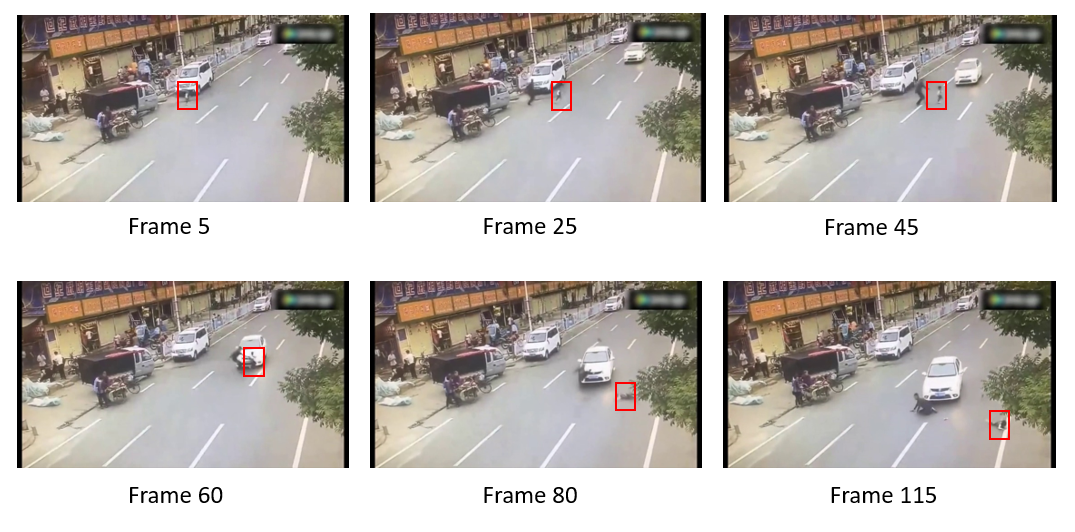}
	\caption{Original Frames in Video 2} \label{fig:video2}
\end{figure}

\begin{figure}[!htbp]
	\centering 
	\includegraphics[scale=0.21]{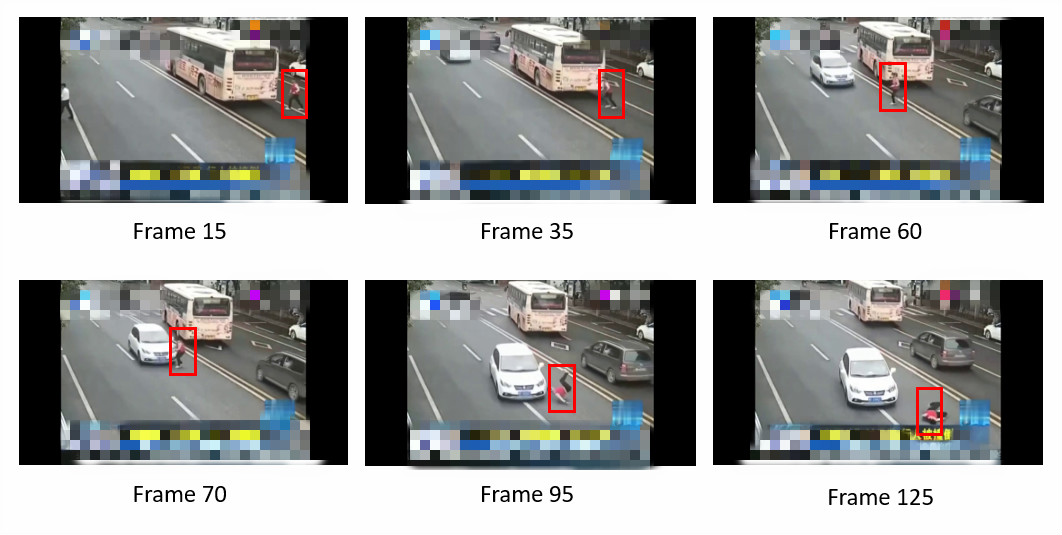}
	\caption{Original Frames in Video 3} \label{fig:video3}
\end{figure}

\begin{figure}[!htbp]
	\centering 
	\includegraphics[scale=0.52]{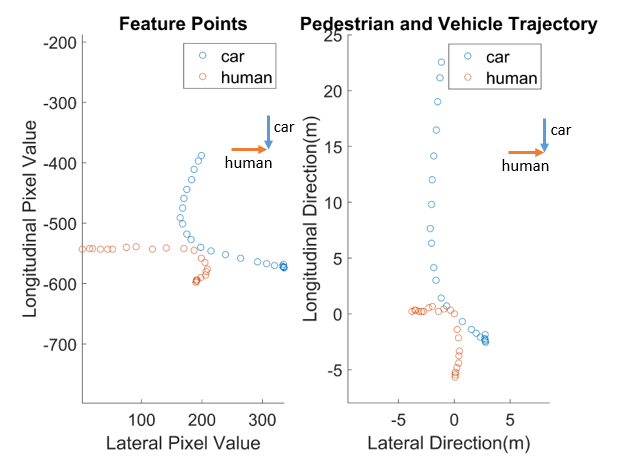}
	\caption{Feature Points and Trajectory of Video1} \label{fig:traj1}
\end{figure}

\begin{figure}[!htbp]
	\centering 
	\includegraphics[scale=0.52]{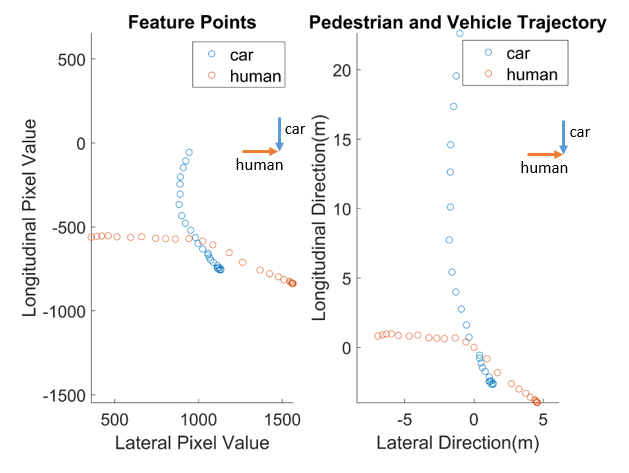}
	\caption{Feature Points and Trajectory of Video2} \label{fig:traj2}
\end{figure}

\begin{figure}[!htbp]
	\centering 
	\includegraphics[scale=0.52]{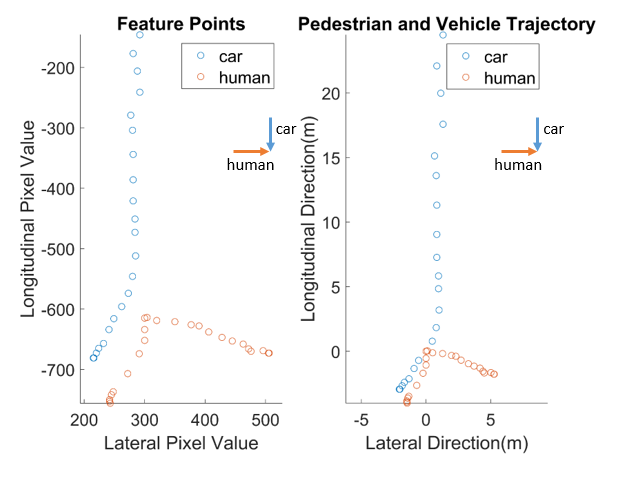}
	\caption{Feature Points and Trajectory of Video3} 
	\label{fig:traj3}
\end{figure}

Fig.~\ref{fig:video1}, Fig.~\ref{fig:video2} and Fig.~\ref{fig:video3} show three original car accident videos\textcolor{black}{, and Fig.~\ref{fig:traj1}, Fig.~\ref{fig:traj2} and Fig.~\ref{fig:traj3} are} the corresponding comparisons between the feature points from the image sequences and the trajectories after scaling. Vehicle and pedestrian movements are in the longitudinal and lateral directions, respectively. The hitting point is set to be $(0,0)$. The object's actual movement trajectory can be estimated from these comparisons, and the objects' trajectories in a traffic incident can be revealed more accurately and completely. 
What's more, given the constant time intervals, the trajectories also approximately indicate the object's velocity and rotation based on the feature point distributions. The points are dense in the low-velocity region and turn to few and scattered while in the high-velocity region. Meanwhile, the vehicle rotation angle can be calculated based on the relative distance of each point.

\begin{figure}[!htbp]
	\centering 
	\includegraphics[scale=0.3]{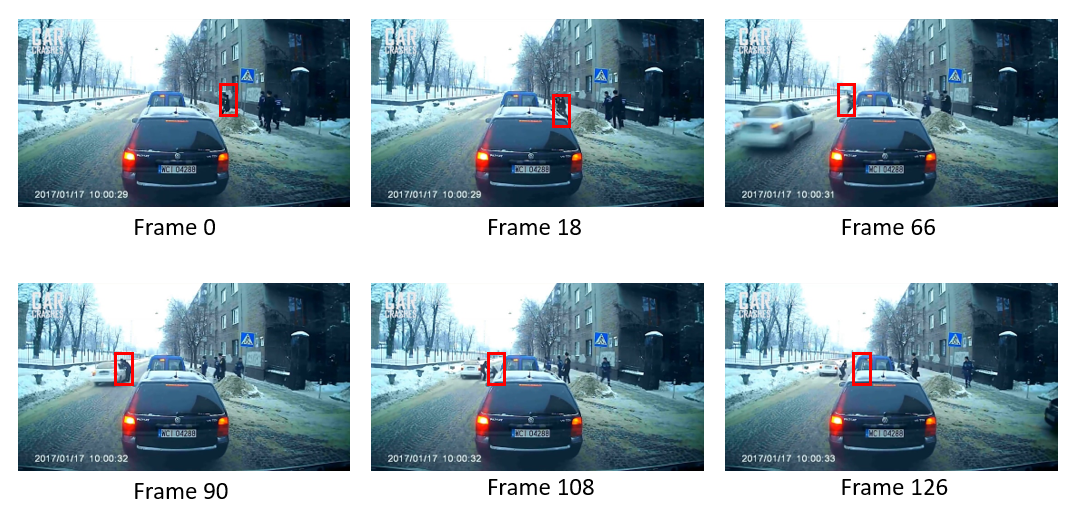}
	\caption{Original Frames in Video4}  \label{fig:video4}
\end{figure}
\begin{figure}
%\vspace{-5pt}
	\centering 
	\includegraphics[scale=0.52]{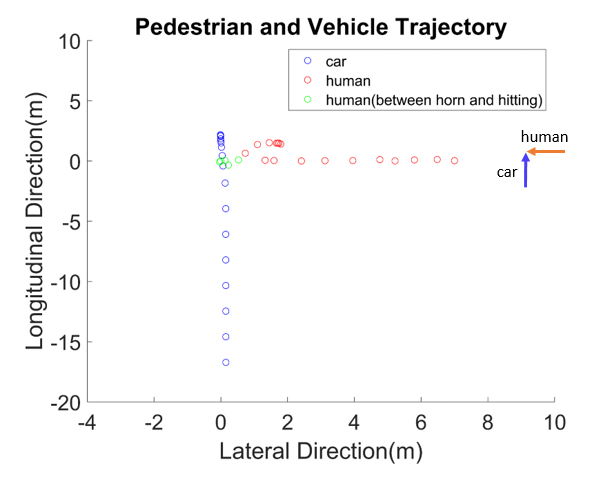}
	\caption{Trajectory of Video4} \label{fig:traj4}
\end{figure}

\vspace{-15pt}
One advantage of collecting the human-vehicle interactive data lies in the fact that the actions other than braking of the vehicles such as the horning and lighting can be captured. Such feature is currently not available in other available open-source data sets. The feature is critically important when studying the human response to the warning actions of vehicles in the actual emergency situations. To capture this feature, we also collect the videos that capture the warning actions of the vehicle. One example is illustrated in Fig.~\ref{fig:video4}, a car accident video captured by a surveillance camera in the street. After applying the perspective transformation and the proposed scaling algorithm, the trajectories of the pedestrian and the vehicle are obtained in Fig.~\ref{fig:traj4}. Considering that the vehicle doesn't show up in the first few image frames, we assume the vehicle's initial movement is straight. For the pedestrian, the points in Fig.~\ref{fig:traj4} not only present the pedestrian displacement, but also indicate human's instinctive reaction in an emergency situation with the warning actions, i.e., horning, from the vehicle. The green points compose a relatively dense region, and this region reveals that the pedestrian's movement slows down after hearing the horn from the vehicle.   

\subsection{Driving Recorder in Car}
\subsubsection{Object Trajectory Reconstruction}
When analyzing the video data recorded by the driving recorder, the scene of the camera is changing with the vehicle movement all the time. Considering the shapes of objects in the scene are magnifying, trajectory reconstruction in the driving recorder employs the perspective transformation in a different way. First of all, operators establish an accident video's last image frame to be the target frame due to the largest shapes of objects. Next, operators select four corner points to form a rectangular object, and the generated code regards these four points that are the target points of perspective transformation as the reference. Ultimately, the perspective transformation is applied on other image frames, and it turns mobile perspective observation to fixed perspective observation. The error caused by the vehicle spacial displacement is eliminated when observing an accident. This method unifies the coordinate system when doing data extraction, and records the pedestrian instinctive reaction details when the pedestrian encounters dangerous.

\vspace{-10pt}
\begin{figure}[!htbp]
	\centering 
	\includegraphics[scale=0.33]{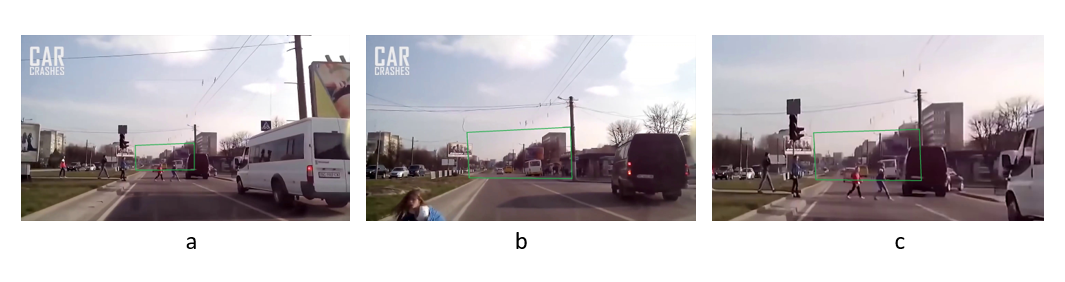}
	\vspace{-5pt}
	\caption{Perspective Transformation Illustration}
	\label{fig:illustration}
\end{figure}
\vspace{-10pt}

The pic $a$ in Fig.~\ref{fig:illustration} is an original video's first image frame, and pic $b$ is the scene after the pedestrian falls to the ground in the same video. Since characterizing the pedestrian trajectory before and after the collision is desirable, the scene after the collision is used as a target picture for the perspective transformation. A quadrilateral composes of the utility pole and the road width is taken as a transformation reference in the picture. The pic $c$ is the picture formed by the perspective transformation according to the target points in pic $b$. Comparing pic $b$, the pedestrian displacement is on the same horizontal line. Hence, by utilizing the perspective transformation, the pedestrian displacement is constrained within a fixed and perpendicular coordinate system. This coordinate system reduces spacial error when extracting pedestrian movement data. 

\begin{figure}[!htbp]
	\centering 
	\includegraphics[scale=0.3]{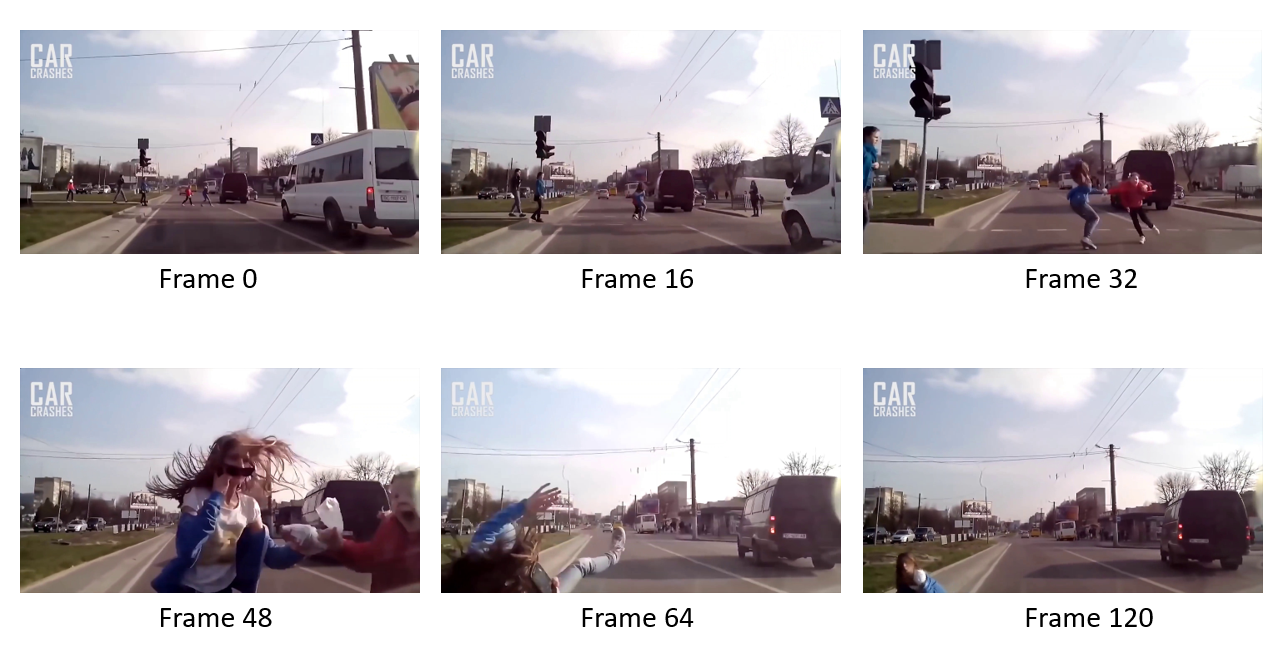}
	\caption{Original Frames in Video5} \label{fig:video5}
\end{figure}

\begin{figure}[!htbp]
%\vspace{-5pt}
	\centering 
	\includegraphics[scale=0.52]{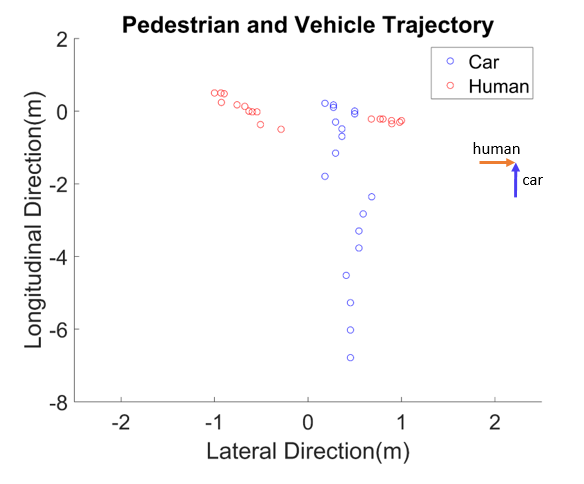}
	\caption{Trajectory of Video5} \label{fig:traj5}
\end{figure}

Fig.~\ref{fig:video5} is an image sequence of the driving recorder in a car. 
The vehicle doesn't show up in the camera and the pixel shapes of objects in the scene are always changing, thus it's difficult to establish a reference to reveal the vehicle's movement. The vehicle's trajectory is roughly shown in Fig.~\ref{fig:traj5}, and the human feature points in Fig.~\ref{fig:traj5} are more dense on the right side of the vehicle trajectory. On the other hand, considering the only object is pedestrian, the camera in-car records more details about pedestrian behavior. The Pedestrian shows hesitation and goes backward to avoid the vehicle in an emergency situation based on the frames in Fig.~\ref{fig:video5}. 

\section{Conclusions}
This paper describes a method to generate trajectory from traffic accident videos. Generated code utilizes image processing to transform perspective, then scales feature points by defining a geometric model. The result shows that pedestrian and vehicle movements are reconstructed accurately. However, we do perspective transformation and mark feature points manually, therefore this extraction approach presents along with slight subjectivity and can’t be a generalized model.
Meanwhile, objects' velocity and vehicle's rotation can be estimated based on the feature point distributions. Given the result, it's beneficial for analyzing driver and pedestrian behavior in a traffic accident. The vehicle rotation angle reveals driver conservative or aggressive rotating behavior, meanwhile, the density of feature points indicates driver braking behavior and pedestrian behavior in an emergency situation. 
Therefore, this method can not only restore a traffic accident scene but also be used to analyze driver and pedestrian behavior to avoid potential traffic incidents.       

\bibliographystyle{IEEEtran}
\bibliography{ref}{}
\end{document}